# How Relevant Are Chess Composition Conventions?

Azlan Iqbal[1]

[1] College of Information Technology, Universiti Tenaga Nasional, Kampus Putrajaya,
Jalan IKRAM-UNITEN, 43000 Kajang, Selangor, Malaysia
azlan@uniten.edu.my
Tel/Fax: 6(03) 89212334

**Abstract.** Composition conventions are guidelines used by human composers in composing chess problems. They are particularly significant in composition tournaments. Examples include, *not having any 'check' in the first move of the solution* and *not 'dressing up' the board with unnecessary pieces*. Conventions are often associated or even directly conflated with the overall aesthetics or beauty of a composition. Using an existing experimentally-validated computational aesthetics model for three-move mate problems, we analyzed sets of computer-generated compositions adhering to at least 2, 3 and 4 comparable conventions to test if simply conforming to *more* conventions had a positive effect on their aesthetics, as is generally believed by human composers. We found slight but statistically significant evidence that it does, but only to a point. We also analyzed human judge scores of 145 three-move mate problems composed by humans to see if they had any positive correlation with the computational aesthetic scores of those problems. We found that they did not. These seemingly conflicting findings suggest two main things. First, the right amount of adherence to composition conventions in a composition has a positive effect on its perceived aesthetics. Second, human judges either do not look at the same conventions related to aesthetics in the model used or emphasize others that have less to do with beauty as perceived by the majority of players, even though they may mistakenly consider their judgements 'beautiful' in the traditional, non-esoteric sense. Human judges may also be relying significantly on personal tastes as we found no correlation between their individual scores either.

## 1   Introduction

A chess problem or composition is a type of puzzle typically created by a human composer using a chess set. It presents potential solvers with a stipulation, e.g. *White to play and mate in 3 moves*, and is usually composed with aesthetics or beauty in mind. Compositions often adhere to many 'composition conventions' as well. Examples include: *possess a solution that is difficult rather than easy*; *contain no unnecessary moves to illustrate a theme*; *have White move first and mate Black*; *have a starting position that absolutely must be possible to achieve in a real game, however improbable*. A more comprehensive list and supporting references are provided in section 3.3.1 of [1]. Composition tournaments or 'tourneys' are at present held all over the world and attract competitors from diverse backgrounds [2].

These conventions exist and are adhered to because they are generally thought to improve the quality or beauty of a composition [1]. They are also useful as a kind of standard so that "*like is compared with like*" [3]. In section 3.2 of [1], a case is made for how not all conventions are prerequisites for beauty. Even so, many composers and players tend to conflate conventions with aesthetics, i.e. a 'good' composition – one that adheres to conventions – is a more *beautiful* one. Award-winning compositions are therefore among the most *beautiful*. In this article, we put this belief to the test as it tends to lead to confusion in the world of composition and how the public understands what they produce.

A review of relevant material relating to computational aesthetics in chess can be found in chapter 2 of [1]. Notably, section 2.4 of [1] explains how modern computer chess problem composition techniques starting in the late 1980s have managed to produce compositions at varying degrees of efficiency and 'quality'. However, the issue of aesthetics and how conventions actually relate to aesthetics is not explored in detail and left largely to the purview of human experts. Included also are comparable works related to Tsume-Shogi, the Japanese equivalent to chess problems. The same chapter illustrates the many problems associated with deriving an aesthetics model from the somewhat vague methods employed by human chess problem judges alone. Our methodology for this research is presented in section 2. In section 3 we explain the experimental setups and results. Section 4 presents a discussion of these results. We conclude the article in section 5 with a summary of the main points and some ideas for future work.

## 2  Methodology

In this research, we used an experimentally-validated computational aesthetics model [4] to evaluate the beauty of three-move mate problems. It has been shown to be able to evaluate and rank aesthetics or beauty in a way that correlates positively and well with *domain-competent human assessment*, i.e. not necessarily 'experts' but also people with sufficient knowledge of the domain to appreciate beauty in it. The model uses formalizations of well-known aesthetic principles and themes in chess in combination with a stochastic approach, i.e. the inclusion of some randomness. All the necessary information regarding its logic, workings and validation can be obtained by the interested reader in [4].

A computer program called CHESTHETICA, which incorporates the model, was used to automatically compose three-move mate problems [5] and evaluate their aesthetics. This was necessary in the first experiment (see section 3.1) – in which we tested the idea that adherence to *more* conventions leads to increased beauty – because *human* compositions tend to contain more variations (alternative lines of play) and variety of conventions than was feasible to calculate manually for each composition. Computer-generated compositions tend to feature just one forced line and fewer, more easily identifiable conventions. The problems used in this research are therefore not of the 'enumerative' kind [6]. The composing module of the program is entirely separate from the aesthetics-evaluating one. Ideally, the latter should be usable to aid the former; however, doing so has proven to be exceedingly challenging. A useful analogy may be how the ability to *rank* beautiful pieces of art does not easily translate into the ability to *create* beautiful pieces of art.

The aesthetics model incorporated into CHESTHETICA assesses primarily 'visual appeal' (see Appendix A of [4] for examples) which is what the majority of chess players and composers with sufficient (not necessarily expert) domain knowledge understand by 'beauty' in the game. Essentially, this includes, for example, tactical maneuvers like sacrifices or combinations that achieve a clear objective such as mate. 'Depth' appeal, on the other hand, relates more toward strategic or long-term maneuvers – perhaps involving many alternative lines of play – amounting to a rather esoteric understanding of the game, furthermore specifically in relation to a particular class of chess composition, e.g. three-movers, endgame studies.

In the second experiment (see section 3.2), human judge scores for human-composed problems were compared against the computer's aesthetic scores to see if there was any good, positive correlation. The underlying idea is that, aside from the slippery concept of 'originality', since human judges tend to emphasize adherence to conventions [1, 3] and consider their judgements pertaining largely to 'beauty' in the sense understood by most players and composers, we would expect that there exists such a correlation with the computer's assessments. Except, of course, in unusual circumstances where there is sufficient compensation in some other aspect of the composition that the judge finds attractive. Together, these two experiments shed some light on the role conventions play in terms of 'beauty' with regard to chess problems and whether human judges are, in fact, scoring beauty as perceived by most players or something else no less relevant to their established art form [7, 8, 9, 10].

## 3  Experimental Setups and Results

### 3.1  Conventions and Aesthetics

For the first experiment, we had CHESTHETICA automatically compose as many three-move mate problems as possible in the time available to us using both 'random' and 'experience' approaches. The 'experience' approach was based on a database of human compositions. In short, pieces are placed at random on the board or based on the probability where they are most likely to be in a chess problem. They are then tested using a chess engine to see if a forced mate exists; see [5] for a more detailed explanation. The 'experience' approach tends to be slightly more effective at composing than the random one and the two are tested here also as an extension of previous work (ibid).

For the first set of composing 'attempts', a filter of two composition conventions was applied so that the resulting compositions would 1) *not be 'cooked'*, and 2) *have no duals in their solution*. A chess problem is said to be cooked when there is a second 'key move' (i.e. first move) not intended by the composer. A solution to a composition is said to contain a 'dual' when White has more than one valid continuation after the key move. For the second set, a filter of three composition conventions was applied so that the resulting compositions would have 1) *no 'check' in the key move*, 2) *no captures in the key move*, and 3) *no key move that restricted the enemy king's movement*. For the third set, a filter of four composition conventions was applied; namely the two conventions from the first set and the first two from the second set.

These conventions were selected because they could be determined with relative certainty and were easier to implement programmatically than others. Based on the literature surveyed (see section 3.3.1 of [1]) there is no particular aesthetic distinction between them or even a strict hierarchy of importance. It is important to note that since thousands of generated compositions needed to be tested for validity, manual determination of conventions was simply not feasible and doing so would have been prone to much human error. Also, the fact that, for example, two conventions were confirmed in the first set and three in the second does not exclude the possibility that more conventions – even those other than CHESTHETICA could detect – were not present, however unlikely. What can be said with some confidence is that the first set contained *at least* two conventions, the second set contained *at least* three, and the third *at least* four.

For the first two sets, there was a total of 120,000 composing attempts run in batches of 1,000 attempts, i.e. where the computer tries to generate a composition that meets all the defined criteria of success. This took approximately 70 days using two standard desktop computers running 24 hours a day. For the third set, several different computers were run simultaneously over a period of approximately 5 months in order to produce the valid compositions. As the number of conventions increases, the efficiency reduces. The composing approach consumes a lot of time primarily because there are far more 'misses' than 'hits' when a chess engine is used to determine if the particular configuration of pieces produced leads to a forced mate.

The computer program used, CHESTHETICA, is also not optimized for this particular composing task. It was designed primarily to evaluate the aesthetics of a move sequence. It is not simply a matter of having more CPU cycles at one's disposal because the approach to automatic composition incorporates many different modules (e.g. random number generation, 'intelligent' piece selection, probability computation, error-correction mechanisms, looping, mate solver) that can take time to produce something, not unlike with human composers. An analogy might be existing chess-playing engines. Simply having more processing power does not necessarily make for a better engine. The quality of the heuristics and other technologies used are also highly relevant.

Table 1 shows the results. Set 3 has no composing attempts and efficiencies listed that can be compared with the other two sets because the attempts were handled differently due to time constraints. Based on past experiments, the efficiencies for the random and experience approaches for set 3 are similar, i.e. between 0.03 to 0.05%. Despite the slightly higher mean composing efficiencies using the 'experience' approach, they were not different to a statistically significant degree from the mean composing efficiencies of the random approach. As anticipated in [5], using conventions as a filter significantly reduces the productivity of the automatic composer.

**Table 1.** Automatic composing results.

|  | Set 1 | | Set 2 | | *Set 3* | |
|---|---|---|---|---|---|---|
|  | **Random** | **Experience** | **Random** | **Experience** | *Random* | *Experience* |
| **Composing Attempts** | 30,000 | | 30,000 | | - | |
| **Conventions Adhered** | 2 | | 3 | | *4* | |
| **Successful Compositions** | 429 | 459 | 303 | 329 | *413* | *297* |
| **Mean Composing Efficiency** | 1.43% | 1.53% | 1.01% | 1.10% | - | - |
| **Total Compositions** | 888 | | 632 | | *710* | |

Table 2 shows the results in terms of aesthetics. The increase of 0.067 in aesthetic value in using 3 conventions instead of 2 was minor but statistically significant; two sample t-test assuming unequal variances: t(1425) = -2.72, P<0.01. The decrease of 0.12 in aesthetic value in using 4 conventions instead of 3 was also minor but statistically significant; two sample t-test assuming equal variances: t(1340) = -4.77, P<0.01. Realistically, we would not usually consider small differences in aesthetic values relevant. However, given that computer-generated compositions were used and an increase of only one convention as a basis of discrimination, we are hesitant to dismiss the findings. On a side note – in relation to an extension of previous research [5] – there was a statistically significant increase in the quality of compositions generated using the 'experience' over 'random' approach for set 1 but not for set 3. For set 2, the *decrease* was not significant.

**Table 2.** Aesthetic scores of the computer-generated compositions.

|  | Set 1 | | Set 2 | | Set 3 | |
|---|---|---|---|---|---|---|
|  | **Random** | **Experience** | **Random** | **Experience** | **Random** | **Experience** |
| **Conventions Adhered** | 2 | | 3 | | 4 | |
| **Mean Aesthetic Score** | 2.167 | 2.241 | 2.307 | 2.240 | 2.148 | 2.158 |
| **Standard Deviation** | 0.48 | 0.50 | 0.46 | 0.45 | 0.473 | 0.458 |
| **Mean Aesthetic Score** | 2.205 | | 2.272 | | 2.152 | |
| **Standard Deviation** | 0.49 | | 0.45 | | 0.47 | |

### 3.2 Human Judge Ratings and Aesthetics

For the second experiment, we looked at the human judge ratings of 145 compositions by human composers. These three-movers were taken from the 'FIDE Album 2001-2003'. Unfortunately, we are unable to make these positions and their scores publicly available even though other researchers may obtain them by purchasing the album [11]. In that system, three judges score each composition on a scale of 0 to 4 and the scores are then summed. The higher the total, the better the composition is considered to be. Details pertaining to judging and selection are available at [12]. Notably, there is nothing explicitly related to aesthetics mentioned. Here is an excerpt.

*"Using a scale of 0 to 4 including half-points, each judge will allocate points to the entries, in accordance with the guidelines shown in Annex 1. The whole scale should be used, but the very highest scores should not occur often. The normal score for a composition good enough for publication in a magazine but without any point of real interest is 1 or 1.5 points. A composition known by the judge to be totally anticipated will attract a score of 0. A composition believed to be unsound but not computer-testable should be given a score nonetheless, since it may turn out to be sound after all. A judge who considers a composition to be incorrect should send his claim and analysis to the director together with his score."* [12]

"ANNEX 1: MEANING OF THE POINT-SCALE

4: Outstanding: **must** be in the Album
3.5
3: Very good: **ought** to be in the Album
2.5
2: Good: **could** be in the Album
1.5
1: Mediocre: **ought not** to be in the Album
0.5
0: Worthless or completely anticipated: **must not** be in the Album" [12]

A chess composition is said to be 'anticipated' when its theme has already appeared in an earlier problem without the knowledge of the later composer. The board configuration therefore does not have to be exactly the same. The 145 problems from the album were also analyzed using CHESTHETICA three times on a scale of 0 to at most 5. There is actually no hard upper limit but no three-mover has ever been found to exceed 5. Due to its stochastic element, the computational aesthetics model may deliver a slightly different score the second or third time it is used to evaluate a composition. Ideally, an average score is used if a crisp value is desired. In this case, however, it was considered more suitable that three evaluations of a composition were totaled just like the three human judge scores.

Incidentally, none of the 145 problems from the album had a score of '0' attributed by any judge so we did not have to compensate for the aesthetics model's inability to detect 'lack of originality' by filtering them out, for instance. CHESTHETICA itself is not available to the public but a version of the program that can evaluate and rank three-move problems and endgame studies in terms of aesthetics is available [13], though this version cannot compose chess problems. Table 3 shows an example of how the human judge scores and computer scores were recorded.

**Table 3.** Sample human judge and computer scores.

|  | Human Judge Scores | | | |
| --- | --- | --- | --- | --- |
|  | Judge 1 | Judge 2 | Judge 3 | Total |
| **Composition 1** | 2 | 2.5 | 3 | 7.5 |
| **Composition 2** | 3.5 | 4 | 3.5 | 11 |
|  | Computer Scores | | | |
|  | Round 1 | Round 2 | Round 3 | Total |
| **Composition 1** | 1.679 | 1.699 | 1.639 | 5.0 |
| **Composition 2** | 1.753 | 1.753 | 1.773 | 5.3 |

The actual scores themselves need not use the same scale because the Spearman or rank correlation was applied. For consistency, the computer's evaluations – in total and average – were always rounded to one decimal place to match the format of the human judge scores. Beyond that, the remaining dissimilar precision in both scales (0.1 vs. 0.5) were not arbitrarily adjusted for. We found no correlation (0.00533; two-tailed, significance level of 1%) between the judge *total* scores for the 145 compositions and computer's *total* scores for them. We tested the *mean* judge scores against the *mean* of the computer's scores and still found no correlation (-0.00523; same). In other words, there was absolutely no aesthetic relationship between the human judge scores and the computer's scores. In fact, there was no significant (Pearson) correlation between the scores of judges 1 and 2 ($r = 0.062$), judges 2 and 3 ($r = -0.036$) and judges 1 and 3 ($r = 0.115$). This suggests that even between judges there is little agreement.

## 4  Discussion

In the first experiment which examined the significance of using more conventions to attain greater aesthetic quality (see section 3.1), we found a very small yet statistically significant increase in adhering to one more convention but only in the incremental step from 2 conventions to 3. The opposite effect was found in adhering to 4. Even though no standard distinction or hierarchy of significance is known in conventions, some are clearly more related to aesthetics than others. For instance, *avoid castling moves because it cannot be proved legal* has likely less to do with beauty than say, *no 'check' in the key move*. The five conventions used in the first experiment are probably of the type that is associated more with aesthetics and this is why the results were suggestive of their contribution to beauty.

Human judges, on the other hand, do not usually standardize which conventions they should look for. Assuming they are as objective as humanly possible, they will evaluate or rate compositions by looking at both conventions that are associated with beauty and those that are less so. Not to mention factors that have little to do with anything other judges might consider relevant. Human judges also consider other intangible concepts such as 'originality' and cannot completely ignore their personal tastes. This might explain why, in the second experiment (see section 3.2), we found no correlation between the human judge scores and the computer's. The issue is when the scores or rankings given by these judges are said to be based on "beauty". Beauty, as perceived by the majority of chess players and composers, is unlikely what these judges are mainly evaluating. This is not to say that human judges have no right to use the word 'beauty' but this research would suggest that that sort of beauty is actually a combination of other things, including personal taste, that is less likely to be understood by the public.

Despite that, the evaluations of these judges are no less viable than they were before because it simply means that 'winning' compositions are not necessarily the most *beautiful*, as the term is commonly understood. There are special things about award-winning compositions that few others outside the domain of expert composition would fully understand, but 'beauty' or aesthetics as evaluated by the model plays only a small part in it. Figs. 1 and 2 show the highest-scoring and lowest-scoring three-movers, respectively, from the collection of computer-generated compositions used in the first experiment and the collection of 145 compositions by human composers used in the second experiment. Only the main lines are shown.

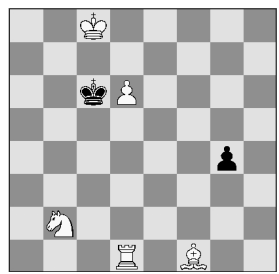     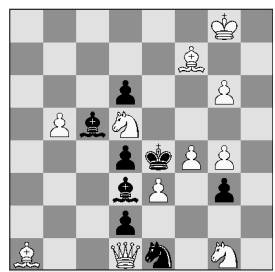

1. Na4 g3 2. d7 g2 3. d8=N#    1. Be6 Bc2 2. Qxd2 dxe3 3. Bf5#
(a)                             (b)

**Fig. 1.** The highest-scoring computer composition (a) and judge-rated problem (b).

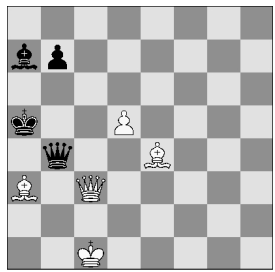     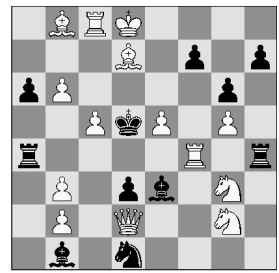

1. Qxb4+ Ka6 2. Bd3+ b5 3. Qxb5#    1. Rf6 Rad4 2. Qc3 Rc4 3. Rd6#
(a)                                  (b)

**Fig. 2.** The lowest-scoring computer composition (a) and judge-rated problem (b).

Readers with sufficient knowledge of chess should be able to form an opinion as to how much human judges are factoring in what we understand by 'beauty' in the game. Notably, beauty in the judge-rated problems appears to be more complicated and understood properly by relatively few (depth appeal) whereas beauty in the computer-generated compositions appears to be more easily perceived and understood by the majority of players and even composers (visual appeal). Readers with no understanding of the game might reach the same conclusion based simply on what they can see from the positions above.

As for the 'random' versus 'experience' approaches (see section 3.1), the results suggest that the latter is no worse, aesthetically, than the former but in compositions filtered using fewer conventions, it can be better. This is not inconsistent with previous findings [5].

## 5 Conclusions

The results of this research suggest that adhering to more conventions, to a point, increases the perceived aesthetic value of a chess problem and that human judges are probably not factoring this sort of (visual) beauty into their rankings or assessments. These findings are important because adherence to more conventions is often confused with increased aesthetics, and because the term 'beauty' is often bandied about in the world of chess composition when it carries a somewhat different meaning outside that esoteric domain.

Aside from certain conventions, the assessment criteria for chess problems are vague and dependent largely on the judges themselves. It is not uncommon for human judges to also be in disagreement with each other about the merits of a composition. Even so, their assessments do result in what we call 'depth appeal' (see Appendix A of [4] for an example) which is sort of a deep appreciation of the theme and variations of play that relatively few with domain competence (e.g. a club player or casual composer) could understand properly. Such appreciation usually occurs after careful study of the problem and is not immediately obvious.

If the aim of experienced composers is greater publicity and accessibility to their art form [3], then more emphasis on 'visual appeal' would be prudent in tourneys and published compositions. However, if this is considered unsuitable, then at least a clarification of what they are really looking at when evaluating chess problems would be wise as the term 'beauty' can be quite misleading, especially outside specialized composing circles. Figs. 1 and 2 above perhaps illustrate the contrast between what the majority of chess players and casual composers understand by 'beauty' and what judges of tourneys do.

Further work in this area may involve examining the use of even more conventions to see if the downtrend continues or improves beyond the use of just 3. Experimentation in this regard is likely to be more difficult because automatic chess problem composition will require an exponentially longer amount of time. Human judge evaluations of other types of compositions (e.g. endgame studies) can be examined as well to see if there is any correlation with aesthetics based on the model used.

# References

1. Iqbal, M.A.M.: A Discrete Computational Aesthetics Model for a Zero-sum Perfect Information Game. Ph.D Thesis, Faculty of Computer Science and Information Technology, University of Malaya, Kuala Lumpur, Malaysia, (2008), http://metalab.uniten.edu.my/ ~azlan/Research/pdfs/phd_thesis_azlan.pdf
2. Giddins, S.: Problems, Problems, Problems. ChessBase News, 16 April, http://www.chessbase.com/newsdetail.asp? newsid=6261 (2010)
3. Albrecht, H.: How Should the Role of a (Chess) Tourney Judge Be Interpreted? The Problemist, July, 217–218 (1993). Originally published as Über Die Auffassung Des Richteramtes In Problemturnieren, Problem, January, 107–109 (1959)
4. Iqbal, A., Heijden, H. van der., Guid, M., Makhmali, A.: Evaluating the Aesthetics of Endgame Studies: A Computational Model of Human Aesthetic Perception. IEEE Transactions on Computational Intelligence and AI in Games: Special Issue on Computational Aesthetics in Games, 4(3), 178-191 (2012)
5. Iqbal, A.: Increasing Efficiency and Quality in the Automatic Composition of Three-move Mate Problems. Lecture Notes in Computer Science, 6972, 186–197. Anacleto, J.; Fels, S.; Graham, N.: Kapralos; B., Saif El-Nasr, M.; Stanley, K. (Eds.). 1st Edition, XVI. Springer. ISBN 978-3-642-24499-5 (2011)
6. Elkies, N. D.: New Directions in Enumerative Chess Problems, the Electronic Journal of Combinatorics, 11(2), pp. 1-14 (2005)
7. Osborne, H.: Notes on the Aesthetics of Chess and the Concept of Intellectual Beauty. British Journal of Aesthetics 4: 160 – 163 (1964)
8. Humble, P. N.: Chess as an Art Form. British Journal of Aesthetics, 33, 59-66 (1993)
9. Troyer, J. G.: Truth and Beauty: The Aesthetics of Chess Problems. In Haller (ed.), Aesthetics (Holder-Pichler-Tempsky, Vienna): 126-30 (1983)
10. Walls, B. P.: Beautiful Mates: Applying Principles of Beauty to Computer Chess Heuristics. Dissertation.com, 1st Edition (1997)
11. Fougiaxis, H., Harkola, H.: World Federation for Chess Composition, FIDE Albums, http://www.saunalahti.fi/~stniekat/pccc/fa.htm, June (2013)
12. Fougiaxis, H., Harkola, H.: FIDE Album Instructions, http://www.saunalahti.fi/~stniekat/pccc/fainstr.htm, January (2013)
13. Iqbal, A., Heijden, H. van der., Guid, M., Makhmali, A.: A Computer Program to Identify Beauty in Problems and Studies (What Makes Problems and Studies Beautiful? A Computer Program Takes a Look). ChessBase News, Hamburg, Germany, 15 December (2012), http://en.chessbase.com/home/TabId/211/PostId/4008602

# Acknowledgement

This research is sponsored in part by the Ministry of Science, Technology and Innovation (MOSTI) in Malaysia under their eScienceFund research grant (01-02-03-SF0240).